\documentclass{article}
\usepackage{graphicx} 
\usepackage[hidelinks]{hyperref}
\usepackage[a4paper, total={6in, 8in}, left=3.5cm, right=3.5cm]{geometry}
\usepackage[backend=biber, style=apa, uniquelist=false]{biblatex}
\usepackage[british]{babel}
\usepackage{url}
\usepackage[hang,flushmargin]{footmisc}
\usepackage{hyperref}
\hypersetup{colorlinks = true, linkcolor = black, urlcolor = gray, citecolor = black, anchorcolor = black}
\usepackage{booktabs}
\usepackage{amsfonts}
\usepackage{nicefrac}
\usepackage{xcolor}
\usepackage{tabularx}
\usepackage{array}
\usepackage{acro}
\usepackage{csquotes}
\usepackage{amsmath}
\usepackage{titlesec}
\usepackage{caption}
\usepackage{changepage}
\usepackage{orcidlink}

\titleformat*{\section}{\normalsize\bfseries}
\titleformat*{\subsection}{\normalsize\bfseries}

\titlespacing*{\section}
  {0pt}{15pt}{0pt}

\newcommand{\affilorcid}[2]{%
$^{#1,\orcidlink{#2}}$%
}

\title{Approaches to Analysing Historical Newspapers Using LLMs}

\author{
Filip Dobranić,\affilorcid{1}{0000-0003-3211-9586}
Tina Munda,\affilorcid{2}{0009-0001-1152-7823}
Oliver Pejić,\affilorcid{1}{0000-0001-8430-7584}
Vojko Gorjanc,\affilorcid{1,2}{0000-0001-9523-347X}
\\
Uroš Šmajdek,\affilorcid{3}{0000-0002-8127-7700},
David Bordon, \affilorcid{2}{0009-0005-4269-552X},
Jakob Lenardič, \affilorcid{1}{0000-0002-4659-3388},
Tjaša Konovšek, \affilorcid{1}{0000-0001-8872-692X},\\
Andrej Pančur, \affilorcid{1}{0000-0001-6143-6877},
Kristina Pahor de Maiti Tekavčič, \affilorcid{1,2}{0000-0002-5678-0881},
Ciril Bohak,\affilorcid{3}{0000-0002-9015-2897}
Darja Fišer\affilorcid{1}{0000-0002-9956-1689}
\vspace{0.5em}
\\
$^{1}$ Institute of Contemporary History, Privoz 11, SI-1000 Ljubljana\\
$^{2}$ Faculty of Arts, University of Ljubljana, Aškerčeva 2, SI-1000 Ljubljana\\
$^{3}$ Faculty of Computer Science, University of Ljubljana, Večna pot 113, SI-1000 Ljubljana
}

\date{}
                 
\begin{document}
\maketitle

\vspace{1.0cm}

\section{Executive Summary}
This study presents a computational analysis of the Slovene historical newspapers \textit{Slovenec} and \textit{Slovenski narod} from the sPeriodika corpus, combining topic modelling, large language model (LLM)-based aspect-level sentiment analysis, entity-graph visualisation, and qualitative discourse analysis to examine how collective identities, political orientations, and national belonging were represented in public discourse at the turn of the twentieth century. Using BERTopic, we identify major thematic patterns and show both shared concerns and clear ideological differences between the two newspapers, reflecting their conservative-Catholic and liberal-progressive orientations. We further evaluate four instruction-following LLMs for targeted sentiment classification in OCR-degraded historical Slovene and select the Slovene-adapted GaMS3-12B-Instruct model as the most suitable for large-scale application, while also documenting important limitations, particularly its stronger performance on neutral sentiment than on positive or negative sentiment. Applied at dataset scale, the model reveals meaningful variation in the portrayal of collective identities, with some groups appearing predominantly in neutral descriptive contexts and others more often in evaluative or conflict-related discourse. We then create NER graphs to explore the relationships between collective identities and places. We apply a mixed methods approach to analyse the named entity graphs, combining quantitative network analysis with critical discourse analysis. The investigation focuses on the emergence and development of intertwined historical political and socionomic identities. Overall, the study demonstrates the value of combining scalable computational methods with critical interpretation to support digital humanities research on noisy historical newspaper data.

\section{Background and Objectives}
Research in historical newspapers has increasingly integrated topic modelling into collaborative digital humanities workflows, bringing together humanities scholars, computer scientists, and information specialists \parencite{oberbichler2022integrated,villamor2023promise}. \textcite{murugaraj2025mining} evaluated topic modelling approaches for newspaper archives, comparing traditional probabilistic Latent Dirichlet Allocation (LDA), matrix factorization-based Non-negative Matrix Factorization (NMF), and neural-based models such as BERTopic \parencite{grootendorst2022bertopicneuraltopicmodeling}. Their findings demonstrate that BERTopic outperforms classical models in all tested aspects, particularly in contextual sensitivity and thematic coherence.

Sentiment analysis has been a longstanding research area, evolving from lexicon-based approaches \parencite{hu2004mining} to machine-learning-based ones \parencite{devlin2019bert,cortes1995support,pang2002thumbs}. In a comprehensive experiment testing the capabilities of LLMs in performing various sentiment analysis tasks, \textcite{Zhang2024-sentiment} highlight the strengths and limitations of LLMs: LLMs excel in simpler tasks, such as binary or trinary sentiment classification, even in zero-shot and few-shot settings, often matching or surpassing fine-tuned smaller language models. This makes them particularly effective when training resources are limited. Slovene sentiment research has also expanded into specialised domains such as finance, where recent benchmarking work evaluates LLMs for target-based financial sentiment in news \parencite{Muhammad2025}. In contrast to these largely contemporary, clean-text settings, the present study targets OCR-degraded historical newspapers and frames the task as aspect-level targeted sentiment classification without supervised fine-tuning.

In the context of digital humanities, graph visualisations have become a common tool for supporting distant reading approaches, while also augmenting close reading by highlighting areas of interest that may warrant closer inspection \parencite{Janicke2015,Janicke2017}. In particular, graphs have proven effective for exploring relationships between named entities extracted from textual corpora. Systems such as NEREx \parencite{ElAssady2017} link entity relationships directly to conversational utterances, enabling users to verify inferred connections in their textual context. Similarly, \textcite{Tamper2023} focus on the exploration of large biographical networks enriched with entity attributes. Building on these approaches, \textcite{Kusnick2024} extend support to multiple entity types and provide a robust query system for filtering and exploring knowledge graphs.

The objective of this study is to combine contemporary computational methods, including the use of LLMs, with established qualitative analysis methods from the humanities. We use both distant and close reading when investigating complex phenomena in heterogeneous historical corpora. By combining the computational strength of analyses with the nuanced insights of traditional humanistic inquiry, it thus enables a novel understanding of historical narratives, patterns, and contexts.

\section{Datasets}
\subsection{Corpus}
The corpus for this study was extracted from the 1 bn word corpus \textit{sPeriodika} of Slovene periodicals published between 1771 and 1914 \parencite{dobranic2024lightweight}. We selected \textit{Slovenec} [Slovene] (1873–1945) and \textit{Slovenski narod} [The Slovene Nation] (1868–1943), two most prominent and widely read Slovene-language political dailies during the turn of the century, catering to readerships with opposing political views but comparable in the number and volume of published issues (see Table \ref{tab:corpus}). While \textit{Slovenec} served as the leading voice of Slovene political Catholicism, \textit{Slovenski narod} was closely aligned with liberal-progressive politics. The principal difference between the two newspapers lay in their stance towards secularism and the Church's role in society. \textit{Slovenec} campaigned for preserving the independence of the Church and its supremacy in education and civic life. While it also supported Slovene nationalist demands, its primary discursive enemy was liberalism, which it often equated with German politics. Conversely, \textit{Slovenski narod}'s rhetoric placed greater emphasis on Slovene nationalism and heavily criticised the unequal status of Austria's non-dominant nationalities. Its main discursive enemies were German nationalism and Slovene political Catholicism, and its core readership consisted of educated professionals as well as wealthier peasants \parencite{amon2011slovensko}.

\begin{table}[h]
\centering
\caption{Corpus size.}

\begin{tabular}{l|ccc}
\toprule
& \textbf{Tokens} & \textbf{Paragraphs} & \textbf{Issues} \\
\midrule
Slovenski narod & 183,294,799 & 4,404,531 & 14,039 \\
Slovenec & 137,506,802 & 3,158,842 & 10,897 \\
\bottomrule
\end{tabular}
\label{tab:corpus}
\end{table}

\subsection{Lexicon of Collective Identities}
A central concept in this study are collective identities which refer to: (i) ethno-national denominations (e.g., Španec [Spaniard], nemški [German], Jud [Jew]), (ii) regional or provincial identities (e.g., Istran [Istrian], Moravka [Moravian]), and (iii) other geography-based identities (e.g., evropski [European]).

We compiled a lexicon of collective-identity-denoting lemmas in three steps. For nominal references, we manually inspected the lemma frequency lists of \textit{Slovenec} and \textit{Slovenski narod}. For adjectival references, lemma candidates were extracted automatically based on Slovene derivational suffixes (e.g., -ski, -ški, including historical orthographic variants such as -zki and -žki). Candidates were filtered by frequency (minimum 90 occurrences) and subsequently manually inspected. The validated lists from both newspapers were merged, deduplicated, and consolidated into a single lexicon used for dataset-wide extraction.

To enable unified analysis, adjectival lemmas were mapped to their corresponding nominal identity categories (e.g., nemški [German] $\rightarrow$ Nemci [Germans], italijanski [Italian] $\rightarrow$ Italijani [Italians]), allowing nominal and adjectival realizations to be grouped under shared identity labels during aggregation.

The lexicon is available for download through the GitLab repository at \url{https://dihur.si/muki/llm4dh/}.

\subsection{Entity-based Sentiment Annotation Evaluation Set}
To evaluate model performance, we constructed a manually annotated dataset of collective-identity mentions sampled from \textit{Slovenec}, \textit{Slovenski narod} and \textit{Slovenka}. Although analyses in this study are performed only on \textit{Slovenec} and \textit{Slovenski narod}, examples from the \textit{Slovenka} newspaper were included in the evaluation set to ensure greater diversity of linguistic contexts during model evaluation and to avoid overfitting to the discourse patterns of only two newspapers. In total, 400 mentions were sampled using stratified sampling to ensure balanced representation across newspapers and grammatical realisations (50\% nominal and 50\% adjectival mentions per newspaper). Sampling was random within these constraints, with an additional frequency cap to avoid dominance of highly frequent identities: if a single identity exceeded 15\% of a newspaper-specific subset, excess instances were replaced via random resampling. The resulting dataset therefore covers a broad range of identity categories across newspapers and grammatical forms.

Annotation was performed at the mention level. Annotators consulted the sentence containing the target mention and, when necessary, up to two preceding and two following sentences to determine sentiment. Mentions were labeled as positive (+), negative (-), or neutral (0), with neutral assigned when no explicit or clearly implied evaluation was present. Mentions whose sentiment could not be determined due to preprocessing errors were marked as \textit{unknown} and excluded. The final evaluation set therefore contains 371 annotated mentions. In addition to sentiment, annotators recorded the referential type for adjectival mentions. Nominal identity expressions were treated as group references by default (e.g., Nemci [Germans]), whereas adjectival expressions were classified according to the semantic type of the modified noun: adjectives modifying collective actors (e.g., German army) were labelled as group references, while adjectives modifying inanimate or abstract entities (e.g., Slovene bread, German politics) were labelled as non-group references. This variable was later used to analyse differences in model performance across referential contexts.

The evaluation set is available for download through the GitLab repository at \url{https://dihur.si/muki/llm4dh/}.

\section{Methods}
\subsection{Topic Modelling}
We use BERTopic \parencite{grootendorst2022bertopicneuraltopicmodeling} to model the topics in each of the periodicals on individual paragraphs. Each of the periodicals is modelled individually using the same set of parameters and random seeds for the UMAP \parencite{mcinnes2018umap} and the topic model with \textit{paraphrase-multilingual-MiniLM-L12-v2} \parencite{reimers_sentence-bert_2019} used for embeddings.

The model produced 40,340 topics for \textit{Slovenec} and 49,537 for \textit{Slovenski narod}. For our analysis, we needed a more manageable number of less fine-grained thematic clusters that would be easier to compare across the periodicals and across time. We tested automated hierarchical clustering, but it produced unsatisfactory results. Instead, we grouped the 500 most-represented topics (comprising roughly two thirds of all paragraphs) into 20 manually curated themes. We excluded out of scope paragraphs (uncategorised topics, textual fragments, garbled text due to OCR errors), which comprise the theme \textit{Outliers and noise} and represent 85.1\% of paragraphs in \textit{Slovenec}, and 88.4\% of paragraphs in \textit{Slovenski narod}. The final set of themes used in our analysis and their size is presented in Table \ref{tab:themes}.

\subsection{Aspect-level Sentiment Analysis with LLM}
The task is formulated as aspect-level sentiment classification with an LLM. For each extracted collective-identity mention, the model predicts the sentiment expressed toward that specific mention within its local context.

Each instance is represented as a structured entry containing:
\begin{enumerate}
    \item a unique mention identifier,
    \item the target identity mention explicitly marked using XML-style tags, and
    \item a context window consisting of the sentence containing the mention together with the two preceding and two following sentences.
\end{enumerate}

Given this input, the model assigns one of three labels to the marked mention: positive (+), negative (-), or neutral (0). Sentiment classification was performed using four instruction-following LLMs representing a range of architectures and training regimes: \textbf{GaMS3-12B-Instruct}, \textbf{Gemma-3-12B-IT}, \textbf{Llama-3.1-8B-Instruct}, and \textbf{DeepSeek-R1-Distill-Qwen-7B}. These models were selected to compare a language-adapted model against widely used general-purpose instruction-following models of comparable scale.

Models were prompted in Slovene with explicit instructions to perform targeted sentiment classification of the marked mention only. Since sentiment predictions were generated at the level of individual collective-identity mentions, predictions from the selected model (GaMS3) were aggregated by collective identity and newspaper for the purposes of the analysis. For each identity within each newspaper, we computed the counts and relative proportions of +, -, and 0 labels.

The full prompt is available for download through the GitLab repository at \url{https://dihur.si/muki/llm4dh/}.

\subsection{Entity Graph Visualisations}
\label{subsec:entytigraphvisualisations}
After annotating all paragraphs with the themes they belong to, we find and extract collective identities (based on our lexicon), the sentiment towards those entities (determined by GaMS), and any locations (based on annotations already present in \textit{sPeriodika}) featured in those paragraphs. These are the nodes used to build our undirected graph. The nodes are connected based on co-occurences of the following pairs: theme-identity, theme-location, identity-location and identity-sentiment. The edge weight is the number of co-occurences.

When visualising the network, colours and node shapes determine node type (see figure captions). Edge thickness is determined by the square root of edge weight, i.e. the number of co-occurrences between two connected nodes. Node size is constant for all but the collective identity node type, where it reflects the relative share of non-neutral sentiment associated with that identity. It is calculated as the proportion of positive and negative occurrences combined, specifically as: 
$$1 - \frac{\text{number of occurrences with neutral sentiment}}{\text{number of occurrences}}$$

\subsection{Discourse Analysis}
National identity is closely connected to geographical locations such as countries, regions, or cities. Media discourse contributes to the construction of these identities by linking collective actors to places and by producing narratives of belonging and exclusion. \textcite{anderson1983imagined} highlights how nations are discursively constructed through shared representations of territory and culture. Such representations often involve processes of othering, in which groups are described through oppositional categories of “us” and “them” \parencite{said1978orientalism, hall1997representation}.

We use a graph representation that connects collective identities, geographical places, and detected sentiment. Selected graph nodes are used to retrieve associated textual segments from the corpus. These segments are categorised according to discourse patterns of national and place-based belonging as well as othering. The analysis is further deepened through close readings of the selected textual segments, allowing for a qualitative interpretation of the discourse and a critical evaluation of the identified patterns.

\section{Results}
\subsection{Distribution of Themes and Collective Identities}
\begin{table}[h]
\centering
\caption{List of themes with absolute and relative paragraph counts, relative share of collective identity mentions and no. of different identities mentioned.}
\begin{tabular}{m{3.5cm}|cccc|cccc}
    \toprule
    \textbf{Theme} & \multicolumn{4}{c|}{\textbf{Slovenec}} & \multicolumn{4}{c}{\textbf{Slovenski Narod}} \\
    & Abs. &  Rel. & Rel. & No. of & Abs. &  Rel. & Rel. & No. of \\
    & paras & paras & ident. & ident. & paras & paras & ident. & ident. \\
    \midrule
    Advertisements and announcements & 15521 & 5.22\% & 4.67\% & 28 &23522 & 6.97\% & 4.56\% & 25 \\
    Art and culture & 6270 & 2.11\% & 4.75\% & 23 & 7730 & 2.29\% & 4.64\% & 23 \\
    Countries and nationalities & 38879 & 13.08\% & 5.96\% & 28 & 35150 & 10.41\% & 4.8\% & 27 \\
    Criminality and natural disasters & 13401 & 4.51\% & 4.65\% & 27 & 14012 & 4.15\% & 4.0\% & 3 \\
    Education & 11845 & 3.98\% & 4.72\% & 12 & 10665 & 3.16\% & 4.64\% & 26 \\
    Family & 2008 & 0.68\% & 4.57\% & 28 & 661 & 0.2\% & 4.81\% & 28 \\
    Finance & 13235 & 4.45\% & 4.98\% & 26 & 26874 & 7.96\% & 4.49\% & 27 \\
    Food production & 3171 & 1.07\% & 5.65\% & 20 & 9652 & 2.86\% & 4.57\% & 11 \\
    Health and mortality & 42490 & 14.29\% & 4.23\% & 25 & 41000 & 12.14\% & 4.63\% & 28 \\
    Infrastructure & N/A &  &  &  & 2387 & 0.71\% & 4.73\% & 27 \\
    Narrative & 11920 & 4.02\% & 5.08\% & 28 & 12304 & 3.64\% & 4.53\% & 28 \\
    Nature and weather & 15960 & 5.37\% & 4.37\% & 13 & 25391 & 7.52\% & 4.5\% & 13 \\
    Newspaper publishing & 2293 & 0.77\% & 4.6\% & 13 & 4482 & 1.33\% & 4.28\% & 19 \\
    Non-Slovenian text & 1243 & 0.42\% & 4.62\% & 11 & N/A &  &  &  \\
    Occupations & 3002 & 1.01\% & 4.61\% & 18 & 4702 & 1.39\% & 4.9\% & 19 \\
    Paratext & 64207 & 21.6\% & 5.20\% & 26 & 60794 & 18.0\% & 4.29\% & 27 \\
    Political life & 14822 & 4.99\% & 4.99\% & 23 & 26530 & 7.86\% & 4.61\% & 28 \\
    Religious practice & 14877 & 5.0\% & 5.41\% & 26 & 7242 & 2.14\% & 4.61\% & 25 \\
    Slovenstvo & 2206 & 0.74\% & 4.38\% & 27 & N/A &  &  &  \\
    Social life & 1163 & 0.39\% & 4\% & 3 & 3776 & 1.12\% & 5.22\% & 18 \\
    State administration & 13144 & 4.42\% & 5.92\% & 29 & 12271 & 3.63\% & 4.85\% & 20 \\
    Travel and communications & 5661 & 1.9\% & 4.14\% & 15 & 8533 & 2.53\% & 4.6\% & 28 \\
    \midrule
    Total (all themes) & 297318 & 100\% &  &  & 337933 & 100\% &  &  \\
    \bottomrule
    \end{tabular}
\label{tab:themes}
\end{table}

As shown in Figure \ref{tab:themes}, the biggest themes in both periodicals are Paratext (\textit{Slovenec}: 21.6\% / \textit{Slovenski narod}: 18\%), Health \& mortality (14.29\% / 12.14\%), and Countries \& nationalities (13.08\% / 10.41\%). Themes with the highest share of collective identity mentions are Countries \& nationalities (5.96\%), State administration (5.92\%), and Religious practice (5.41\%) for \textit{Slovenec}, and Social life (5.22\%), Occupations (4.9\%), and State administration (5.92\%) for \textit{Slovenski narod}. Themes with the highest number of different identity mentions are State administration (29), Advertisements \& announcements (28) and Countries \& nationalities (28) for \textit{Slovenec} while Travel and communications (28), Political life (28), Health \& mortality (28) are the most diverse identity-mention-wise in \textit{Slovenski narod}. When looking at the least diverse topics identity-wise, we see Social life (3), Non-Slovene text (11), and Education (12) in \textit{Slovenec} and Criminality \& natural disasters (3), Food production (11), and Nature \& weather (13) in \textit{Slovenski narod}.

For more details on thematic analysis, including diachronic, see (Dobranić et al. 2026).

\subsection{Performance of LLMs for Aspect-based Sentiment Annotation}

\begin{table}[h]
\centering
\caption{Cross-model performance on mention-level sentiment classification (N=371). M-F1 denotes macro-averaged F1; W-F1 denotes support-weighted F1. Class-specific scores are reported as F1\_-, F1\_0, F1\_+, corresponding to NEG, NEU, and POS sentiment labels.}
\begin{tabular}{lcccccc}
\toprule
Model & Acc & M-F1 & W-F1 & F1$-$ & F1$_0$ & F1$+$ \\
\midrule
GaMS3 & \textbf{0.695} & 0.538 & \textbf{0.708} & \textbf{0.488} & \textbf{0.794} & 0.343 \\
Gemma3 & 0.679 & \textbf{0.555} & 0.701 & 0.477 & 0.776 & 0.411 \\
LLaMA 3.1 & 0.485 & 0.459 & 0.515 & 0.406 & 0.545 & \textbf{0.427} \\
DeepSeek-R1 & 0.253 & 0.244 & 0.260 & 0.269 & 0.267 & 0.197 \\
\bottomrule
\end{tabular}
\label{tab:modelsperf}
\end{table}

We evaluated the performance of four cutting-edge instruction-following LLMs: \textbf{GaMS3-12B-Instruct}, \textbf{Gemma-3-12B-IT}, \textbf{Llama-3.1-8B-Instruct}, and \textbf{DeepSeek-R1-Distill-Qwen-7B} in a few shot setting, using the same prompt template, decoding parameters, and manually annotated evaluation sample (N=371 mentions). 

As shown in Table~\ref{tab:modelsperf}, GaMS achieves the highest overall accuracy (0.695) and the highest weighted F1 (0.708), indicating the strongest performance under class imbalance. Gemma attains the highest macro-averaged F1 (0.555), suggesting slightly more balanced performance across sentiment classes.

Given these results, subsequent dataset-level analyses are based on GaMS3-12B-Instruct. Although Gemma yields a slightly higher macro-averaged F1, GaMS provides the highest overall accuracy and weighted F1, as well as the strongest performance on neutral detection, which dominates the historical dataset distribution. This makes GaMS the more suitable model for large-scale aggregation.

\begin{table}[h]
\centering
\caption{Performance of GaMS3 on the manually annotated sample for mention-level sentiment classification for the full dataset (Total). Metrics are: precision (P), recall (R), F1-score (F1). Macro and weighted averages are computed across classes.}
\begin{tabular}{l|cccc}
\toprule
\textbf{Class} & \textbf{P} & \textbf{R} & \textbf{F1} & \textbf{Support} \\
\midrule
Positive (+) & 0.400 & 0.300 & 0.343 & 40 \\
Neutral (0) & 0.858 & 0.739 & 0.794 & 287 \\
Negative (-) & 0.351 & 0.750 & 0.478 & 44 \\
\midrule
\textbf{Macro avg} & 0.536 & 0.596 & 0.538 & 371 \\
\textbf{Weighted avg} & 0.749 & 0.692 & 0.708 & 371 \\
\bottomrule
\end{tabular}
\label{tab:gamsperf}
\end{table}

Table~\ref{tab:gamsperf} summarizes the performance of GaMS3-12B-Instruct on the manually annotated evaluation set. GaMS achieves an overall accuracy of 0.695, with a weighted F1 of 0.708 and a macro-averaged F1 of 0.538. The gap between weighted and macro-F1 reflects class imbalance and uneven performance across sentiment categories.

GaMS detects neutral sentiment most reliably (F1=0.794, P=0.858, R=0.739; support=287). Both precision and recall are comparatively high, indicating stable behavior with relatively few false positives and false negatives in neutral contexts. Negative sentiment reaches an F1 of 0.478 (P=0.351, R=0.750; support=44). The substantially higher recall than precision indicates that the model captures most annotated negative instances but over-assigns negative labels in some neutral or positive contexts. Positive sentiment proves more challenging (F1=0.343, P=0.400, R=0.300; support=40). The relatively low recall suggests that a considerable proportion of positive instances are not recognised and are instead classified as neutral or negative. Compared to the negative class, the model appears more conservative in assigning positive sentiment.

Taken together, the results show a clear asymmetry in class behaviour. Neutral sentiment is the most stable category. Negative sentiment is detected with high sensitivity but less selectively, while positive sentiment is under-detected. For downstream aggregation, this implies that dataset-level summaries may under-represent positive evaluations than negative ones, while neutral proportions remain comparatively robust.

For more details on benchmarking LLMs for aspect-based sentiment annotation in historical newspapers, see Munda et al. (2026).

\subsubsection{GaMS Performance by Grammatical Category}
We also analysed GaMS’s performance separately for nominal identity mentions (e.g., Nemci [Germans], Slovenci [Slovenes]) and adjectival modifiers (e.g., nemški [German], slovenski [Slovene]) to assess whether grammatical form affects classification behaviour.

Performance is weaker for noun mentions than for adjectives. In the nominal subset (N=180), overall accuracy and weighted F1 reach 0.633 and 0.630, compared to 0.749 and 0.777 for adjectives. Neutral sentiment is still the most reliably identified category among nouns (F1=0.742), and negative sentiment is often detected, but with lower precision than recall, indicating a tendency to over-assign negative labels in ambiguous cases. Positive sentiment is especially difficult for nouns: although it is present in the data, the model rarely identifies it correctly–with recall reaching only 0.100 and precision 0.222–and instead tends to default to neutral or negative predictions.

The adjectival subset (N=191) yields substantially better overall results. Neutral sentiment again performs best (F1=0.839), and positive sentiment is detected much more successfully than in the nominal subset (F1=0.488). Negative sentiment, however, shows a different error pattern here: recall reaches 1.00, but precision remains low (0.310), indicating that the model captures nearly all negative adjectival instances at the cost of over-predicting negativity in some neutral contexts.

Overall, these results show that grammatical form affects not only overall performance, but also the balance between precision and recall across sentiment classes. Nominal mentions are associated mainly with missed positive evaluations, whereas adjectival mentions are more prone to over-attributing negative sentiment. This matters for dataset-level analysis, because aggregated identity-level sentiment distributions combine both grammatical realizations and therefore inherit both types of error.

\subsubsection{GaMS Performance by Referential Type}
We next assess whether GaMS’s performance varies according to referential type, distinguishing between group-referential mentions (direct references to collective actors, e.g., Nemci [Germans], češki narod [Czech people]) and non-group mentions (typically adjectival nationality markers modifying inanimate heads, e.g., nemška politika [German politics]).

Performance is weaker for group-referential mentions than for non-group mentions. In the group subset (N=245), accuracy and weighted F1 reach 0.645 and 0.660, compared to 0.786 and 0.802 in the non-group subset (N=126). Neutral sentiment is again the most reliably identified category in both subsets. In group contexts, however, negative sentiment is detected with higher recall than precision, indicating some over-assignment of negative labels, while positive sentiment remains difficult to detect (F1=0.286).

Performance improves in the non-group subset. Neutral sentiment is identified very reliably (F1=0.867), and positive sentiment is also detected more successfully than in the group subset (F1=0.476). Negative sentiment reaches perfect recall in this subset, but this result should be interpreted cautiously because the number of negative instances is very small (6).

Overall, these results suggest that direct group-referential mentions are more difficult for the model than non-group contexts. In group contexts, positive evaluations are more often missed, while negative sentiment is detected more readily. This matters for dataset-level aggregation, as summaries of sentiment toward collective actors may underestimate positive evaluations more than negative ones.

\subsubsection{Use Case: Sentiment Distribution by Collective Identity and Newspaper}
\begin{figure}[h]
    \centering
    \includegraphics[width=0.6\columnwidth]{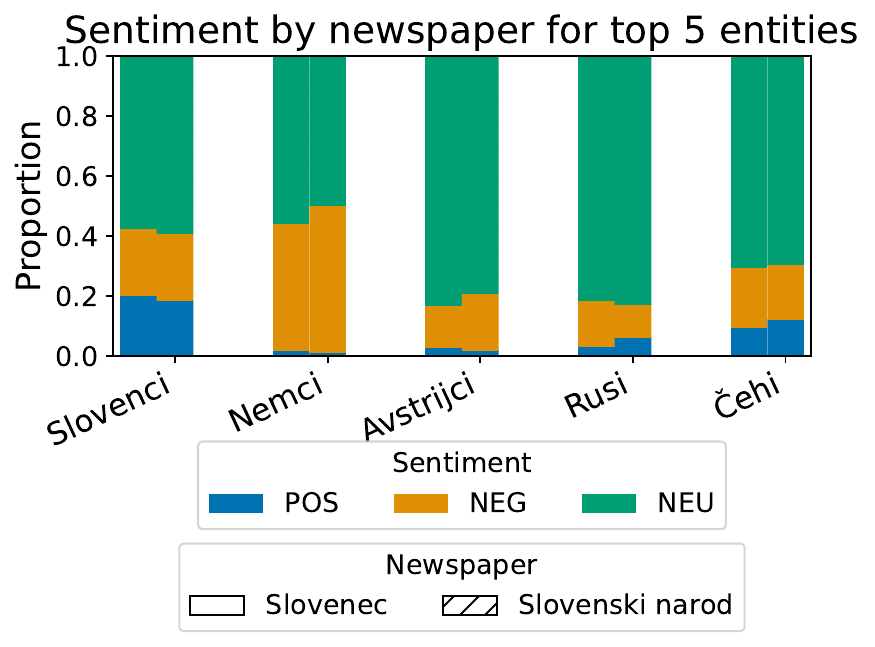}
    \caption{Sentiment class composition (+/-/0) for the five most frequent collective identities in the dataset. For each identity, two stacked bars show the predicted class proportions in \textit{Slovenec}, and \textit{Slovenski narod}.}
    \label{fig:sentiment-top5-composition}
\end{figure}

Following the evaluation and diagnostic analysis above, we applied GaMS3-12B-Instruct to all identity mentions in the dataset and aggregated predicted sentiment labels by identity and newspaper. Figure~\ref{fig:sentiment-top5-composition} shows the proportional distribution of +, -, and 0 sentiment predictions for the five most frequently mentioned collective identities in the dataset.

The distributions differ across identity categories. Nemci [Germans] show a consistently higher proportion of negative predictions in both newspapers, with relatively smaller positive shares. In contrast, references to Slovenci [Slovenes] display a more balanced or mixed distribution, including a visibly larger proportion of positive predictions. Identities such as Avstrijci [Austrians] and Rusi [Russians] are dominated by neutral predictions across newspapers, with only limited positive or negative shares. The distribution for Čehi [Czechs] appears more mixed, with moderate levels of both positive and negative labels depending on the newspaper.

These patterns suggest that the model differentiates between identity categories rather than assigning sentiment uniformly. On the one hand, some of our preliminary findings largely align with mainstream historiographical narratives surrounding turn-of-the-century Slovene history. For example, the overwhelmingly negative sentiment that both newspapers show towards Germans is indicative of contemporary Slovene-German nationalist political conflict \parencite{maribor_svoji_nodate}. Likewise, the predominantly neutral sentiment expressed towards Austrians is expected given that Slovene political discourse of the time was mostly supportive of the Austrian state and Austrian political identity \parencite{luthar2008}.

Conversely, the results that we have gained by analysing the lemma Čehi demonstrates some of the interpretative complexities faced when compiling data using such models. While our interpretation remains speculative, we assume that the negative sentiment in the two journals were connected to the intensity of German-Czech nationalist conflict in the multiethnic crownland of Bohemia [\textit{Češka}] - a province that is otherwise referred to using the same adjective [\textit{češki}] as the Czech nation \parencite{KrenHeumos1996}.

Figure~\ref{fig:sentiment-top5-composition} demonstrates that mention-level predictions can be aggregated at dataset scale while preserving identity-specific variation, provided that model performance characteristics are taken into account.

\subsubsection{Use Case: Most Neutral and Non-neutral Collective Identities}
Since the selected model is substantially more reliable in assigning neutral sentiment than positive or negative sentiment, we next examine identities with the highest proportion of neutral sentiment, as well as identities with the highest proportion of non-neutral sentiment (positive and negative combined). To ensure meaningful comparisons between the newspapers, the analysis was restricted to identities with at least 50 mentions in both \textit{Slovenec} and \textit{Slovenski narod}.

\begin{figure}[h]
    \centering
    \includegraphics[width=0.8\linewidth]{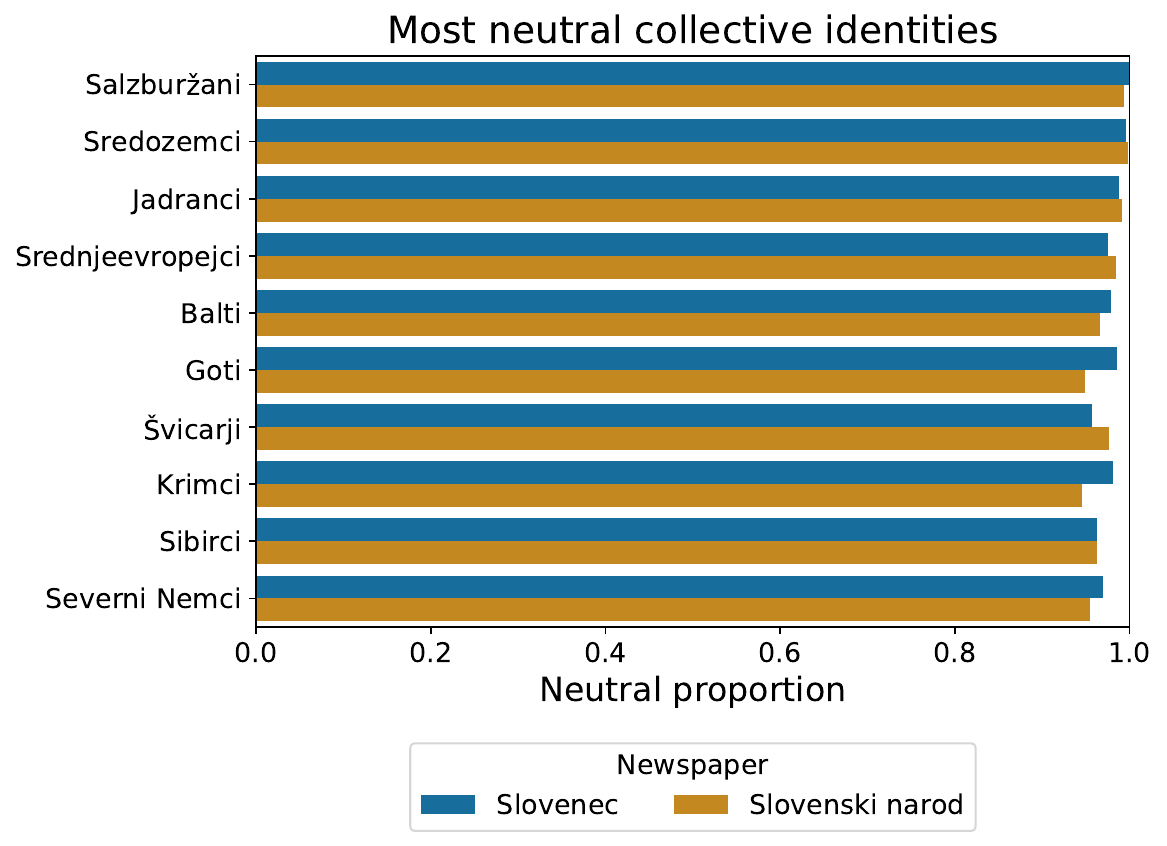}
    \caption{Most neutral collective identities. Top 10 collective identities with the highest proportion of neutral sentiment in \textit{Slovenec} and \textit{Slovenski narod}. Only identities with min. 50 mentions in each newspaper are included. Identities are ordered by the mean neutral proportion across the two newspapers. Bars represent the proportion of mentions classified as neutral by the model.}
    \label{fig:mostneutralidentities}
\end{figure}

Figure~\ref{fig:mostneutralidentities} shows the collective identities with the highest proportion of attributed neutral sentiment. Several broadly regional groups dominate this category, including \textit{Sredozemci}, \textit{Jadranci}, \textit{Adrijanci} and \textit{Srednjeevropejci}. These identities reach neutral proportions close to 1.0 in both newspapers, indicating that they are predominantly mentioned in descriptive or informational contexts rather than in evaluative discourse. The patterns are highly consistent across the two newspapers, suggesting a shared tendency to treat such collective groups as neutral categories.

\begin{figure}[h]
    \centering
    \includegraphics[width=0.8\linewidth]{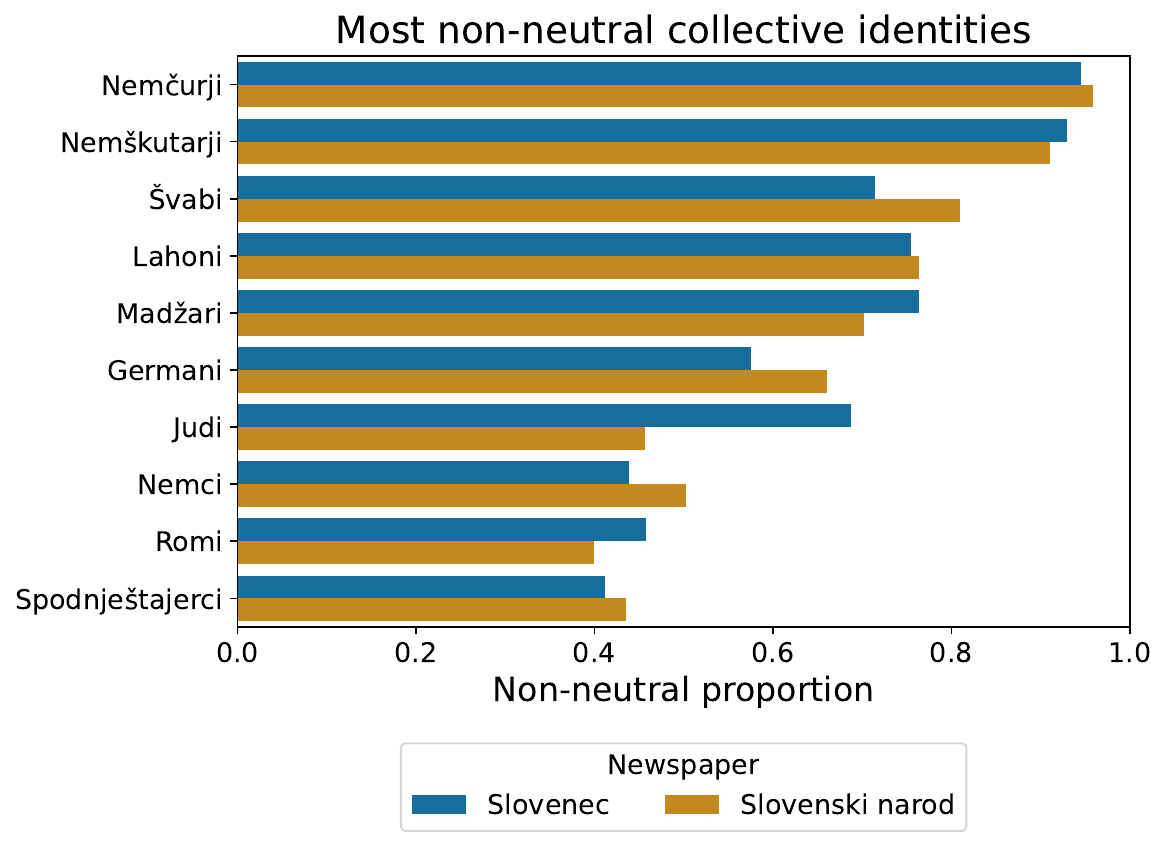}
    \caption{Most non-neutral collective identities. Top 10 collective identities with the highest proportion of non-neutral sentiment (positive or negative) in \textit{Slovenec} and \textit{Slovenski narod}. Only identities with min. 50 mentions in each newspaper are included. Identities are ordered by the mean non-neutral proportion across the two newspapers. Bars represent the combined proportion of positive and negative mentions.}
    \label{fig:mostnonneutralidentities}
\end{figure}

In contrast, Figure~\ref{fig:mostnonneutralidentities} highlights identities with the highest proportion of attributed non-neutral sentiment. These include \textit{Nemčurji}, \textit{Nemškutarji}, and \textit{Švabi}—all pejorative expressions referring to politically pro-German or German-assimilated Slovenes—along with \textit{Lahoni} (pro-Italian/Italianized Slovenes) and \textit{Madžari} (ethnic Hungarians). Many of these identities correspond to historically salient national or political groups in the late nineteenth and early twentieth centuries. Their high non-neutral proportions indicate that mentions frequently occur in evaluative contexts, such as political commentary, conflict reporting, or polemical discourse. While the overall patterns are similar across newspapers, differences in the intensity of non-neutral sentiment for specific identities should be interpreted with caution, since the model tends to over-predict negative sentiment and under-detect positive sentiment.

Taken together, Figures~\ref{fig:mostneutralidentities} and~\ref{fig:mostnonneutralidentities} illustrate a clear contrast in the sentiment distribution associated with different types of collective identities. Broad regional or geographic categories tend to appear in neutral descriptive contexts, whereas politically salient national groups are more often embedded in evaluative discourse. This pattern aligns with expectations for historical newspaper reporting, where geopolitical actors and contested identities are more likely to be discussed in opinionated or conflict-related contexts.

\subsection{Portrayal of Collective Identities}
We compared \textit{Slovenec} and \textit{Slovenski narod} by visualising the network of selected topics, connecting collective identities with locations and sentiment. Specific graph nodes were used to extract related textual paragraphs from the corpus, which were then analysed using critical discourse analysis combined with close reading to interpret common discourse patterns. The distant reading visualisations provide a bird's-eye perspective of network connections, while close reading and discourse analysis allow for a more detailed interpretation of the established links.

\subsubsection{Use Case: Network Analysis}
Due to the size of the corpus (14,926 newspaper issues in total), we chose to inform our close reading efforts by first analysing the collective identities, the sentiment with which the newspapers reported on them, and the locations co-occurring with these identities as networks where nodes are our entities and the edges their co-occurrence.

\begin{figure}
    \centering
    \includegraphics[width=0.7\linewidth]{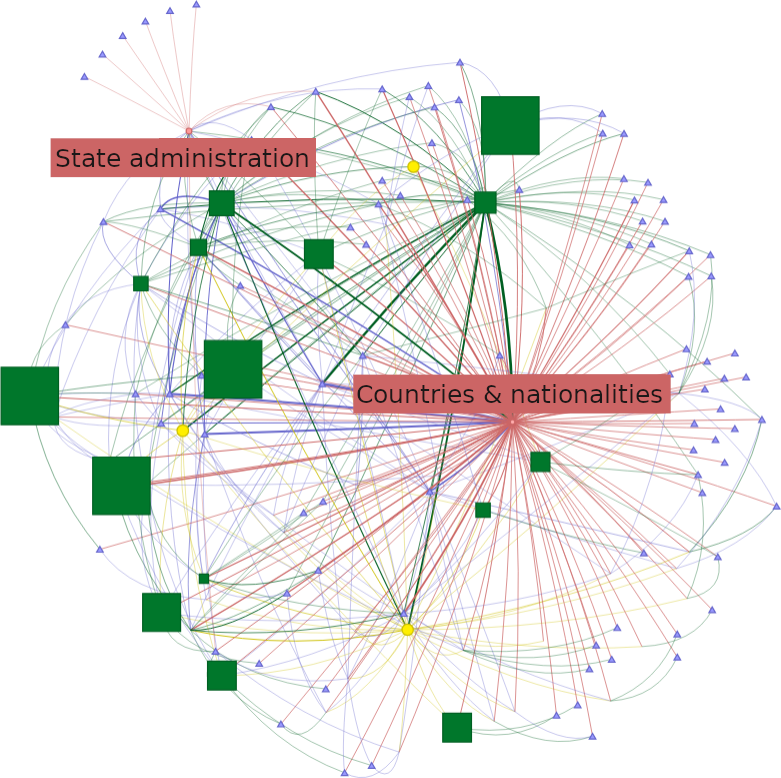}
    \caption{Topic (red), identity (green), sentiment (yellow), and location (purple) graph for \textit{Slovenski narod}. The size of the nodes is the same except for identities, where it correlates to that identity’s relative non-neutral sentiment.}
    \label{fig:slovenskigraph}

    \centering
    \includegraphics[width=0.7\linewidth]{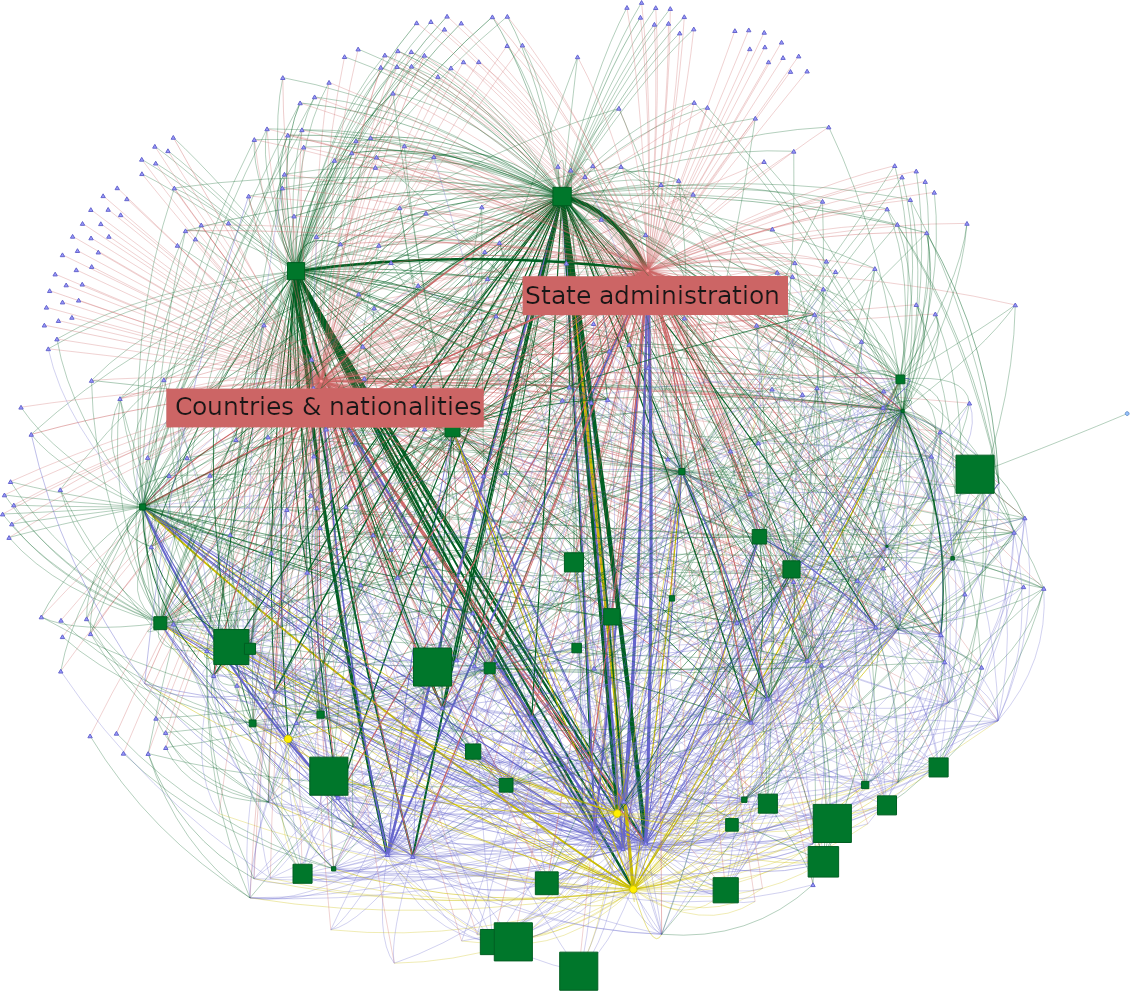}
    \caption{Topic (red), identity (green), sentiment (yellow), and location (purple) graph for \textit{Slovenec}. The size of the nodes is the same except for identities, where it correlates to that identity’s relative non-neutral sentiment.}
    \label{fig:slovenecgraph}
\end{figure}

In Figures~\ref{fig:slovenskigraph} and~\ref{fig:slovenecgraph}, we visualise the network of two topics (Countries and nationalities, State administration), and their relationships with collective identities and locations in the corpus. Comparing the two graphs, we can see a much more diverse and interconnected graph for \textit{Slovenec} as opposed to \textit{Slovenski narod}. Second, the collective identities in the graph for \textit{Slovenski narod} are more often referred to with neutral sentiment, which is indicated by most of them being minuscule in size (see Section \ref{subsec:entytigraphvisualisations}) compared to \textit{Slovenec}.

\begin{figure}
    \centering
    \includegraphics[width=0.8\linewidth]{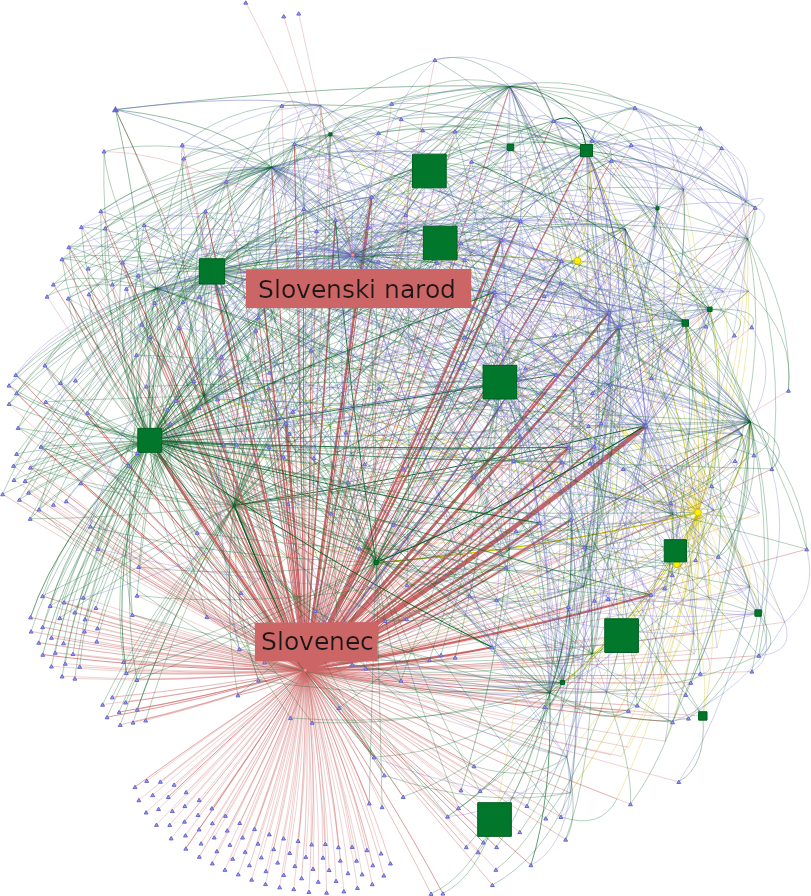}
    \caption{A comparison of \textit{Slovenec} and \textit{Slovenski narod} paragraph co-occurrences for the Advertisements and announcements theme. Topics are shown in red, identities green, sentiment yellow, and locations purple. The size of the nodes is the same except for identities, where it correlates to that identity’s relative non-neutral sentiment.}
    \label{fig:adsgraph}
\end{figure}

Figure~\ref{fig:adsgraph} directly compares a single topic (Advertisements and announcements) in the two newspapers. Here, \textit{Slovenec} shows both a more diverse set of co-occurrences with collective identities and a higher rate of co-occurrence than \textit{Slovenski narod}. The network further shows that \textit{Slovenec} includes a larger number of locations that do not appear in \textit{Slovenski narod}, although many of these are relatively local addresses. \textit{Slovenski narod}, on the other hand, features England, which does not appear in \textit{Slovenec}.

Based on these visualisation, we identified cases for close reading (see Section \ref{subsection:cda}). Our analysis focuses on discursive strategies used to construct a sense of Slovene identity in each of these newspapers.

\subsubsection{Use Case: Critical Discourse Analysis}
\label{subsection:cda}
A \textbf{cross-thematic discourse analysis} with close reading reveals a deliberate use of various strategies to construct a strong sense of Slovene national identity. These strategies include othering, positive self-presentation, and references to historical struggles. One of the most prominent strategies is \textit{othering}, particularly in relation to neighbouring identities or those with whom Slovenes coexisted within the former Austro-Hungarian Monarchy—namely Germans, Italians, Hungarians, and Croats. Croats are presented in a positive context as a brotherly nation with a similar fate in relation to \textit{others}, as exemplified in the statement \textit{Veseli nas, da naši bratje Hrvati tudi na tem polji napredujejo in da se ne pustijo od „privandranih" nemških kričačev komandirati}. (We are pleased that our brothers, the Croats, are also making progress in this field and that they do not allow themselves to be commanded by “newcomer” German instigators). Germans, rather than Italians or Hungarians, are most often positioned as the \textit{others} in relation to Slovene identity. This othering extends beyond nationality to ideological positions, as seen in phrases like \textit{slovenskemu narodu škodljiv tabor — nemški ali klerikalni} (a camp harmful to the Slovene nation—the German or conservative one), and in the use of negative descriptions such as \textit{nemškutarska nesramnost in predrznost} (Germanophile insolence and arrogance).

A key distinction between the two thematic areas analysed—Countries and Nationalities and State Administration—lies in their respective focus. The Countries and Nationalities texts emphasize cultural and historical narratives of belonging, while the State Administration texts focus on governance and autonomy. In the latter, attention is drawn to the need for the presence of Slovenes in public life and the public use of the language, as highlighted in the statement \textit{priporočamo, naj preskrbe /.../ razen nemških in laških tudi slovensko nupise} (We recommend that, in addition to German and Italian, /.../ they also provide Slovene inscriptions). Furthermore, the right of Slovenes to self-regulate their public life is emphasised, as expressed in \textit{Slovenci sami vejo utrditi svoje javno življenje tako, kakor jim najbolje ugaja} (The Slovenes themselves know how to strengthen their public life in the way that suits them best.), reinforcing a sense of self-determination.

A \textbf{cross-newspaper comparison} regarding the reporting on the same group identities—namely Germans, Italians, Hungarians, and Croats—in the context of Advertisements and Announcements reveals a significant difference in the presence of mentions of nationalities. This is primarily due to textual genre differences between the newspapers. In \textit{Slovenec}, the dominant genre containing references to nationalities is the obituary. In this context, non-Slovene nationalities are typically mentioned, thus emphasising the \textit{other}.

At the same time, the context of the obituary also reveals differences in the sentiment toward these nationalities. Mentions of Germans are neutral, such as \textit{nemški pridigar in katehet} (German preacher and catechist) or \textit{nemški pesnik} (German poet). In contrast, mentions of Croats include both neutral and distinctly positive evaluations, such as \textit{vrlega hrvatskega rodoljuba in prvoboritelja} (noble Croatian patriot and trailblazer) or \textit{odličen hrvaški rodoljub} (excellent Croatian patriot). The obituary genre in \textit{Slovenec} appears in both short and longer formats, with detailed presentations of the lives of deceased individuals, providing more opportunities for (positive) evaluations.

A common feature of both newspapers, where nationalities typically appear in the theme of Advertisements and Announcements, is short news items. In \textit{Slovenski narod}, these are explicitly labelled as Drobne novice (Short news), covering topics such as the openings and operations of national institutions (e.g., schools, assemblies) or news from specific national environments, such as Hrvaške vesti (Croatian news). Both genres—obituaries and short news—typically connect collective entities with highly dispersed geographical locations. 

\section{Implications and Conclusion}
This study demonstrates the potential of a mixed-method approach that combines established language technologies and novel LLM-based approaches with qualitative discourse analysis for uncovering sociopolitical and ideological patterns in historical periodicals, thereby offering new insights into the cultural and political landscapes of selected Slovene historical periodicals.

This study evaluated whether instruction-following LLMs can reliably perform targeted, mention-level sentiment classification in OCR-extracted text from historical Slovene newspapers and whether these predictions can support large-scale historical analyses when aggregated across the dataset. We show that the best-performing model GaMS3-12B-Instruct is usable for this task, but its performance is class-dependent and varies across grammatical realization and referential type. We would like to underscore an often overlooked disconnect between technical benchmarks and scholarly needs. For DH workflows, a model's F1 score is ultimately less important than its interpretive validity — its capacities and limitations when used to map the complex, often ambiguous reality of human expression. More broadly, the study provides a benchmark for targeted sentiment classification in OCR-degraded historical Slovene, highlights the necessity of cross-model comparison in DH settings, and offers an empirically grounded assessment of both the reliability and limitations of instruction-following LLMs as analytical instruments in DH research.

We show how network analysis of paragraph-level term co-occurrence can be a productive tool to guide close reading attempts. The mixed-methods approach allowed us to investigate a corpus that is unsuitable for close-reading-based discourse analysis alone.

This research, on the one hand, confirms established historiographical knowledge on topics such as growing nationalist polarisation as well as the impact of key historical events on public discourse. At the same time, it also reveals several novel insights into differences in reporting among the periodicals. These findings underscore the value of computational methods and distant reading in historical and cultural research, particularly in enabling large-scale, data-driven analyses of complex historical texts.

In future work, we plan to further integrate LLMs to perform in-depth analyses of topics and their linguistic framing, investigating how these periodicals constructed narratives around collective identities to achieve a deeper understanding of the cultural and political dynamics of the past. Temporal named entity graphs will be explored to observe the changing prominence and attitudes of collective identities.

\section*{Acknowledgements}

This work was supported by the Slovenian Research and Innovation Agency research programme ``Digital Humanities: resources, tools and methods'' (2022--2027) [grant number P6-0436], the support of the DARIAH-SI research infrastructure, the Slovene Common Language Resources and Technology Infrastructure, CLARIN.SI, and by the project ``Large Language Models for Digital Humanities'' (2024--2027) [grant number GC-0002].

\printbibliography


\end{document}